\documentclass[11pt]{article}
\usepackage{coling2020}
\usepackage{times}
\usepackage{url}
\usepackage{latexsym}

\usepackage{xcolor}
\usepackage{soul}

\usepackage[utf8]{inputenc}
\usepackage[arabic,english]{babel}

\usepackage{graphicx}
\graphicspath{ {./images/} }

\colingfinalcopy 

\title{Arabic Dialect Identification Using BERT-Based Domain Adaptation}

\author{Ahmad Beltagy\qquad Abdelrahman Wael\qquad Omar ElSherief
\\
\\Faculty of Engineering, Alexandria University\\
\{a.beltagy97, abdelrahman.abouelenin\}@gmail.com,
\\C.m\_elshrief@hotmail.com
}

\date{}

\begin{document}
\maketitle
\begin{abstract}
  Arabic is one of the most important and growing languages in the world. With the rise of social media platforms such as Twitter, Arabic spoken dialects have become more in use. In this paper, we describe our approach on the NADI Shared Task 1 that requires us to build a system to differentiate between different 21 Arabic dialects, we introduce a deep learning semi-supervised fashion approach along with pre-processing that was reported on NADI shared Task 1 Corpus. Our system ranks 4th in NADI's shared task competition achieving a 23.09\% F1 macro average score with a simple yet efficient approach to differentiating between 21 Arabic Dialects given tweets.
\end{abstract}

\section{Introduction}
\label{intro}

\blfootnote{
This work is licensed under a Creative Commons  Attribution 4.0 International License. License details: \url{http://creativecommons.org/licenses/by/4.0/}.
}

Arabic dialect classification is a task of identifying the dialect of the writer given an input text. This task has been an active field of research the past few years due to the rise of Arabic corpora which are made available ~\cite{bouamor-etal-2018-madar}. However, NADI's Arabic dialect corpus ~\cite{mageed-etal-2020-nadi} has been quite challenging and intriguing. NADI's corpora has 21 different country dialects from which some of them are quite similar to each other in terms of morphology. Some of the data provided had some English words, others had Quran verses generated from third-party apps which makes it extremely difficult for not just models, but also humans cannot differentiate between the dialects if the tweet is a Quranic verse, since Quran compromises classical Arabic, not dialectal Arabic. This is noise in the dataset that a model can not fix. Also, the imbalance of training data introduced other challenges and difficulties.

Previous work in this task involved the use of traditional ML algorithms, RNN with their variants, and hybrid approaches like in~\cite{salameh2018fine}. However, we were inspired by self-attention technique~\cite{Vaswani2017AttentionIA}.  Due to the huge success of transformers in many classification tasks, we opted for an approach of using pre-trained BERT ~\cite{devlin2018bert} in semi-supervised deep learning as it is proved by ~\newcite{Gururangan2020DontSP} that fine-tuning pre-trained model to a specific domain is an efficient solution. We fine-tuned pre-trained BERT transformer on ``AraBERT"~\cite{antoun2020arabert} Arabic text, which is trained on 23B GB of data, using masked language modeling with Huggingface interface
 ~\cite{Wolf2019HuggingFacesTS}. This process was done after doing data pre-processing and data augmentation in order to alleviate the problems of having class imbalance and lack of training data.

Transfer learning has proven to be an efficient approach in classification tasks, especially when we do not have enough training data. We leveraged the unlabeled data provided by NADI \cite{mageed-etal-2020-nadi} through Twitter API's, for fine-tuning using language modeling to achieve domain adaptation. Where we adapt
pre-trained AraBERT to the domain of tweets. That is because of the similarity of morphological and semantic features that both the unlabeled data and target data share. pre-trained BERT model "AraBERT" was fine-tuned on NADI's unlabeled corpus using masked language model to let it learn better features. These features are going to be used to classify target labeled training examples provided by NADI, which have the same distribution as the unlabeled ones. Our results confirm that such a technique has improved our baseline performance significantly by 3\% F1 macro average score.

The rest of the paper is organized as follows. Section 2 presents the data that was supplied for training our model. Section 3 is a description of our system in detail and what infrastructure we used to produce the results. In Section 4, we present our results using various techniques. Section 5 is a discussion about the task and common errors that occurred. Section 6 concludes the work we have done as well as suggested future work.

\section{Data}
The dataset was one of the factors that made this problem quite challenging. NADI shared task organizers ~\cite{mageed-etal-2020-nadi} provided a corpus of Arabic tweets from Twitter platform having 21 class labels corresponding to countries \{Egypt, Iraq, Saudi Arabia, Algeria, Oman, Emirates, Libya, Syria, Morocco, Yemen, Tunisia, Lebanon, Jordan, Kuwait, Palestine, Qatar, Bahrain, Djibouti, Mauritania, Somalia, and Sudan\}. The corpus is divided into a training set, dev set, and test set to report our final results on. The number of examples in the 3 sets is, 21,000 tweets, 4,957 tweets, and 5,000 tweets respectively. To aid in the training and model building processes, the organizers also provided additional 10 million unlabeled tweets IDs from the same distribution of the labeled tweets, to be obtained using a provided python script.

\begin{figure}[h]
    \centering
    \includegraphics[width=0.8\textwidth]{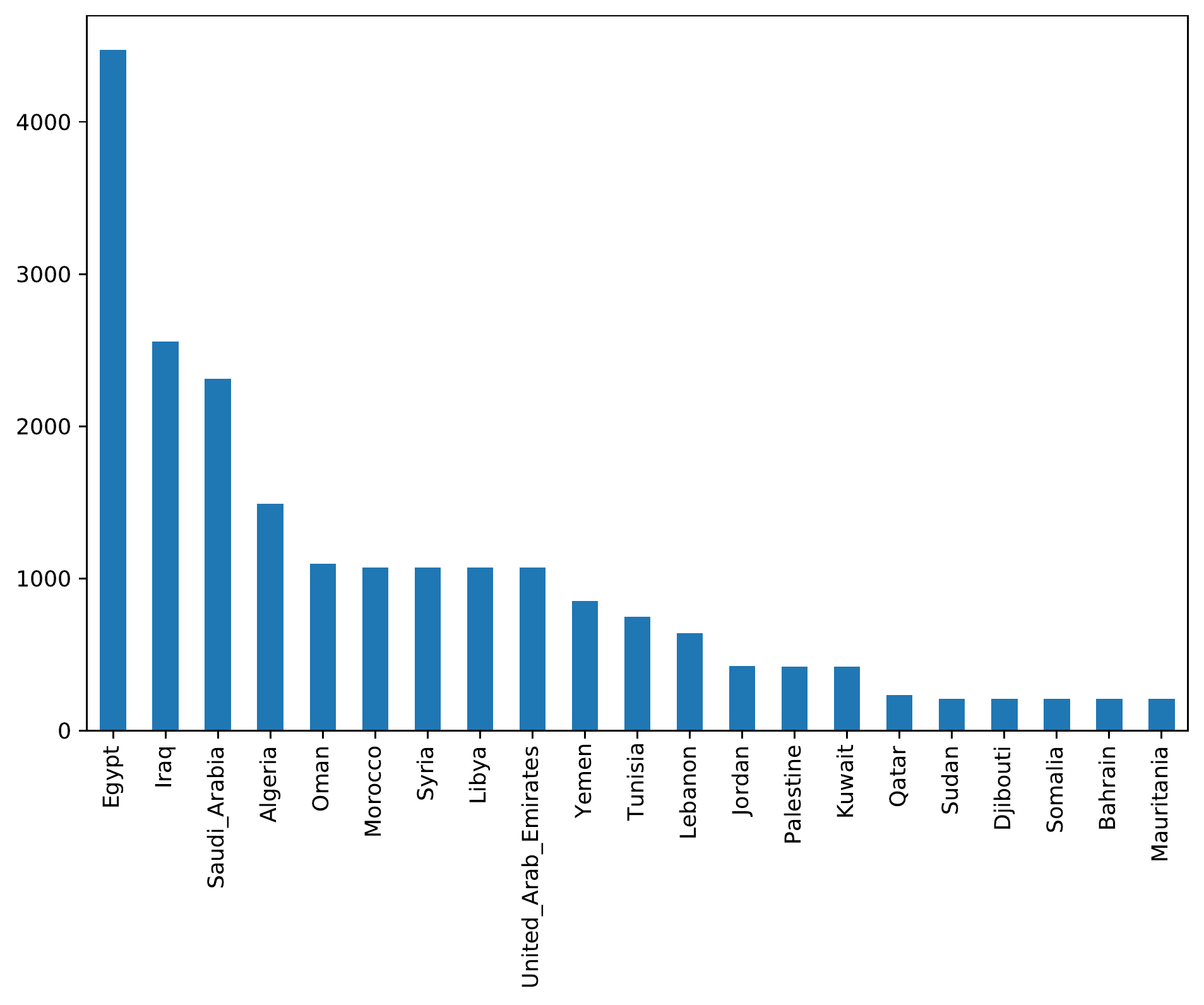}
    \caption{Distribution of 21 class labels across training data}
    \label{fig:dist}
\end{figure}

In Figure \ref{fig:dist}, we show the distribution of each class in the training examples. The figure clearly shows the imbalance of training examples where a class like Egypt has 4,473 examples and a class like Sudan has only 210 examples. It also shows that most of the classes had under 1,000 examples, which was quite challenging to solve.

\subsection{Data Pre-processing}
\label{sect:pre-processing}
We start our pre-processing by cleaning the data and removing all the unnecessary characters. We start the process of data cleaning by removing URLs, punctuation, mentions, email addresses, emojis, and other unknown Unicode characters using regex patterns. Next step we removed all English characters as it won’t help in differentiating between Arabic dialects. Moreover, we used PyArabic library ~\cite{zerrouki2012pyarabic} to strip tashkil from Arabic sentences. We stripped tashkil because we observed the inconsistent use of it and even wrong usage, which it would be a burden rather than an advantage to keep it. At this point, we have a corpus of only Arabic characters. The next step of cleaning is to remove elongations ex: \AR{ والله متععععههههه اهخ لو كملت}. Here we won’t need \AR{متععععههههه} as it represents noise, we remove the repeated characters so the result \AR{متعه}. Next, we deal with the words that have the same semantics but different syntax, those words differ not in the core of the word itself, but in the suffix and prefix, for example, the word \AR{بعرفكيش},  which means {I don't know you}, can be written in different forms with different dialects  \AR{معرفكيش, ما بعرفك, لا أعرفك, مش هعرفك}, so in order to provide more information to our model. We separate the suffix and prefix from the core of the word, for this step we use Farasa Segmenter ~\cite{abdelali-etal-2016-farasa}, which is an Arabic NLP toolkit that serves as a sentence segmentation toolkit. Farasa  takes a word and splits it into suffix and prefix, which completes our data processing step, example: \AR{امين يارب العالمين} this is transformed to  '\AR{ين}'+'\AR{عالم}'+'\AR{امين يارب ال}'

\subsection{Data Augmentation}
\label{sect:upsampling}
Because of the imbalance of data as shown in figure \ref{fig:dist}, some minority classes like \{Lebanon, Jordan, Kuwait, Palestine, Qatar, Bahrain, Djibouti, Mauritania, Somalia and Sudan\} had less than 750 examples. Our proposed solution to this problem is to upsample the minority classes to 750 examples using scikit-learn library ~\cite{scikit-learn}. We chose this specific number after many trials with different numbers. We observed that above 750 examples we do not observe any increase in terms of accuracy. 
Other techniques for data augmentation were suggested in~\cite{fares-etal-2019-arabic,8614166,ibrahim2020alexu}
\section{System}
In this section, we describe our proposed approach used in NADI shared task 1 Corpus submission. All of our experiments were based on AraBERT ~\cite{antoun2020arabert} which is an Arabic version BERT model trained on 23GB of Arabic text with 3B words having vocab size of 64,000. We present the building blocks of our system, then we go over to explain another experiment that we have tried.

\subsection{Tokenization and Encoding}
First, we used the tokenizer corresponding to the model which mainly is used to split the sentence to tokens, example: \AR{خلص يبقى بعرفكيش} this is transformed to  '\AR{ش}','\AR{كي}','\AR{رف}','\AR{بع}','\AR{يبقى}','\AR{خلص}'.
Then we proceed to convert each word to it's appropriate ID and if the tweet is smaller than the expected dimension (64 word), then it's padded with an appropriate token " [PAD] ". If it exceeds it, then it's truncated and a binary vector, also known as a mask, is returned to emphasize if a certain token corresponds to a word or padding token.
Then we feed AraBERT the sequence of IDs along with its corresponding mask vector and its label corresponding to which class of the 21 mentioned in the data section. 

\subsection{Feature Extraction}
AraBERT is originally a pre-trained BERT ~\cite{devlin2018bert} specifically for the Arabic language. It has a vocab size of 64,000 words and uses the same small Bert-Base configuration that has 12 transformer encoders. These transformers encoders learn about features in the input text and output 768 hidden dimensions at the end.
After each sentence is fed to the model, it is converted to a vector representing it using a vocab dictionary and then fed to encoder layers of the model to form an output of (768). 
\subsection{Domain Adaptation Using Fine Tuning}
\label{sect:domain-adaptation}
We were inspired by the power of domain adaptation in various Natural Language Processing tasks. It was proven by ~\newcite{Gururangan2020DontSP} that adapting a large pre-trained model to another domain would gain a lot in terms of accuracy. We started this task by crawling some of the 10M unlabeled tweets using Twitter API python script provided by NADI organizers ~\cite{mageed-etal-2020-nadi}. We only utilized 2M tweets, as we did not observe any gain in accuracy when going beyond 2M tweets. We trained a masked language model using HuggingFace interface ~\cite{Wolf2019HuggingFacesTS}, to let the model learn about semantic features of the new domain 'tweets'. The masked language model was trained using an adaptive learning rate starting with 2e-5 was used and with a masking rate of 15\% of the input text. We trained with only 1 epoch, which took around 20 hours to complete training. The rest of the hyper-parameters were the default ones provided by HuggingFace.

\subsection{Training Classifiers}
We experimented with 2 different settings to classify tweets. First, which was the one we submitted. It relied only on AraBERT to classify the 21 labels, achieving an F1 macro average score of 24.433\% on the dev set. The second Trial was a mix between AraBERT and Naive-Bayes achieving 22.3\% F1 macro average score.

\subsubsection{AraBERT classification head}
After each tweet was encoded into 768 features, a simple BERT classification layer was added after multiple experiments. The layer would take in the 768 features and using a fully connected network it converts them to output a vector of 21 entries. Where each entry uniquely corresponds to one of the 21 Arabic dialect classes. 


\subsubsection{AraBERT with naive-bayes}
\begin{figure}[h]
    \centering
    \includegraphics[width=0.6\textwidth]{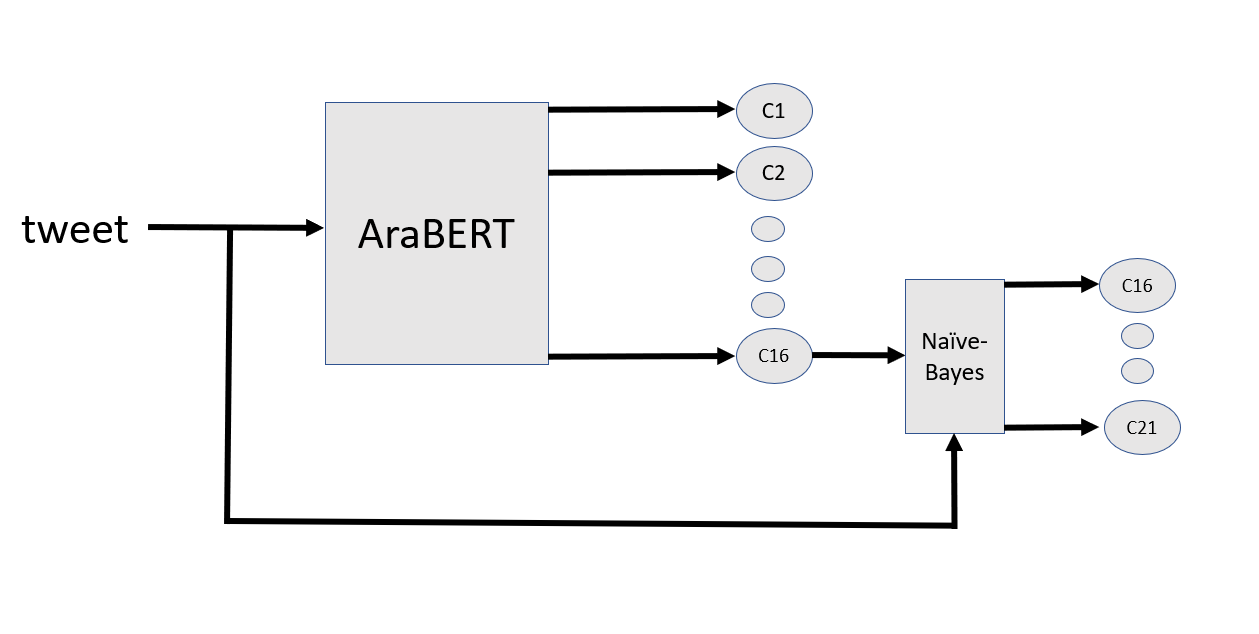}
    \caption{Visual of the classification head used in AraBERT with Naive Bayes. If  AraBERT classifies a certain tweet to belong tominority class then it is passed to Naive-Bayes.}
    \label{fig:araBERT-naive}
\end{figure}

One of the promising models and inspired by ~\cite{Kowsari2017HDLTexHD} was a hierarchical classification model. Having observed that deep learning doesn't perform well with classes with a small number of tweets, we also observed how well naive Bayes performed classifying these classes, so we were motivated to build a hybrid model using AraBERT and naive Bayes.The idea was simple, we used AraBERT to classify majority classes \{Egypt, Iraq, Saudi Arabia, Algeria, Oman, Emirates, Libya, Syria, Morocco, Yemen, Tunisia, Lebanon, Jordan, Kuwait, Palestine\} normally and combining minority classes \{Qatar, Bahrain, Djibouti, Mauritania, Somalia, and Sudan\} into one class \{Not majority\} labeled C16 as shown in figure \ref{fig:araBERT-naive}. If AraBERT classified a certain tweet to belong to this special class then this tweet is passed to naive Bayes to more accurately classify which class from the minority classes does this tweet belongs to.
This model proved capable of achieving an f1 score of 22\% but still falls a bit short than a pure AraBERT.

\section{Results}

\begin{table}[h]
\begin{center}
\begin{tabular}{|l|c|}
\hline \bf Model & \bf Dev-set F1-score  \\ \hline
baseline (AraBERT without augementation and cleaning) & 18.6\%  \\
\hline
AraBERT with cleaning                               & 20.54\%  \\
\hline
AraBERT with cleaning + up sampling                  & 21.33\%  \\
\hline
AraBERT with cleaning + up sampling + MLM fine-tuning  & 24.43\% \\
\hline
AraBERT + Naive-Bayes with cleaning + up sampling      & 22.3\%  \\
\hline
\end{tabular}
\end{center}
\caption{\label{dev-set-results} Our proposed models scores on dev-set. }
\end{table}

We present the results of our experiments on the development set provided by organizers which consisted of 4,957 examples. In Table \ref{dev-set-results} we show our corresponding F1 macro average score on those models. We firstly show that our baseline model which consists of using AraBERT only resulted in an 18.6\% F1 macro average score. After the cleaning process described in data section \ref{sect: pre-processing}, we gained nearly 2\% F1 macro average score which proved our pre-processing was an efficient one. However, we noticed that our classifier is struggling with minority classes as described in the data section \ref{sect: upsampling}, which gave us the idea of upsampling them. And that gained us a further 0.79\% F1 macro average score. After further research, we were fascinated by the power of the fine-tuning approach of training a masked language model to improve classification tasks as described in system section \ref{sect:domain-adaptation} . This technique has given us the biggest improvement of 3.1\% F1 macro average score. This was our submitted model in the task, which ranked 4th with 24.43\% F1 macro average score in dev set and 23.09\% on the test set.

A second experiment that was not submitted was concluded based on the hierarchical model of AraBERT with naive-Bayes. However, this approach had an F1 macro average of 22.3\%, which is still 2\% F1 macro score less than a single transformer model.

\section{Discussion}
It was observed in our experiments, through analysis of data, that some of the tweets had noticeable noise that cannot be solved. Some users used to tweet through third-party apps. We noticed that these tweets were mainly prayers or Quran verses coming from various apps like \{http://d3waapp.org, http://knzmuslim.com, http://Gharedly.com, and http://du3a.org\}. The main problem is that these tweets are given different labels according to the nationality of the user, and in fact they are not even differential by humans, which makes it very hard for a trained model. Another noticeable noise was that some tweets are actually retweets. Where the retweeted content refers to a dialect, but it was given another label because the actual user retweeting is having a different dialect.

In future work, we plan on exploiting hierarchical models having a transformer model, as it's the backbone as we believe that regrouping our classes into different sets, would improve the results.
Also, another promising idea was to build a byte pair encoding tokenizer trained on extracting frequent sub words, which may prove to be helpful especially in dialect classification. We believe that the main differences between dialects are specific sub-words that occurs in certain dialect more than others. We will also experiment with ensemble techniques and how they can fit into our system.

\section{Conclusion}
We introduce a simple efficient single deep neural network model to classify 21 Arabic dialects, which was based on the idea of a fine-tuning approach using a pre-trained transformer model on a similar domain after various cleaning and augmentation methods. We were able to achieve the 4Th best score in NADI shared task 1 competition with an F1 macro average score of 22.03\%.

\bibliographystyle{coling}
\bibliography{coling2020}

\end{document}